\documentclass[prodmode,acmjacm]{acmsmall} 

\usepackage[ruled]{algorithm2e}
\usepackage{mathtools}
\usepackage{subfigure}
\usepackage{tabularx}

\SetAlFnt{\small}
\SetAlCapFnt{\small}
\SetAlCapNameFnt{\small}
\SetAlCapHSkip{0pt}
\IncMargin{-\parindent}

\newcommand{\Ha}{\mathcal{H}}

\newcommand{\norm}[1]{\left\lVert#1\right\rVert}

\usepackage{fancyhdr}
\begin{document}

\markboth{Peeyush K. et al.}{Near Optimal Hamiltonian-Control and Learning via Chattering}

\title{Near Optimal Hamiltonian-Control and Learning via Chattering}
\author{PEEYUSH KUMAR, WOLF KOHN and ZELDA B. ZABINSKY
\affil{ Atigeo, 800 Bellevue Way NE Suite 600, Bellevue, WA}}

\begin{abstract}
Many applications require solving non-linear control problems that are classically not well behaved. This paper develops a simple and efficient chattering algorithm that learns near optimal decision policies through an open-loop feedback strategy.  The optimal control problem reduces to a series of linear optimization programs that can be easily solved to recover a relaxed optimal trajectory. This algorithm is implemented on a real-time enterprise scheduling and control process.
\end{abstract}


\acmformat{Peeyush Kumar, Wolf Kohn and Zelda B. Zabinsky, 2015. Near Optimal Hamiltonian-Control and Learning via Chattering.}
\begin{bottomstuff}
This work is supported by Atigeo, 800 Bellevue Way NE Suite 600, Bellevue, WA 98004.

Author's addresses: Peeyush Kumar {and} Zelda B. Zabinsky, Industrial and Systems Engineering, University of Washington Seattle. Wolf Kohn, Atigeo, 800 Bellevue Way NE Suite 600, Bellevue, WA 98004.
\end{bottomstuff}

\maketitle

\section{Introduction}
Many decision problems involve solving a non-linear control problem that lacks traditional properties such as smooth controls and convex state spaces. This paper develops a chattering algorithm to solve non-linear control problems that are classically not well-behaved. The algorithm works with the Hamiltonian formulation for optimal control problems. This formulation provides many advantages as it can incorporate logic rules (the language of communication in enterprises), can facilitate planning and model adaptation (Hamiltonian describes the time evolution of a system), and scales to big data systems (Hamiltonians are additive for multi-body sytems). In addition, this method is implemented using an open-loop feedback strategy that balances the trade-off between control adaptability and computational efficiency. We demonstrate this method on a real-time enterprise scheduling and control process.  \\

Non-linear control problems are used in diverse applications but they remain challenging to solve computationally. One generally successful method is the State-Dependent Riccati Equation (SDRE) technique \cite{riccatisurvey}. While this method has been used  in many applications, its major limitation is that it requires solving the Riccati equations multiple times. This is generally computationally inefficient. Another recent work in the field solves the non-linear optimal control problem using homotopy perturbation methods \cite{Jajarmi2012759}. This method is computationally efficient as compared to the SDRE technique. However, the approximation methods that are used lack formal analysis on error bounds. It is not well understood how these methods can generalize to ill-behaved, non-differentiable, non-convex systems. \\

Another well studied approach to solving non-linear control problems is dynamic programming and various approximate solutions enable better computational performance \cite{Bertsekas:2007:DPO:1396348,Powell:2007:ADP:1324761}. The reinforcement learning and Bayesian learning literature has many algorithms using dynamic programming for Markov systems \cite{Sutton:1998:IRL:551283,Wiering:2014:RLS:2670001}. These methods are generally adaptive and can solve a wide variety of problems, although their success is largely limited to discrete Markovian systems or discretized approximations to continuous Markovian systems. Dynamic programming approaches do not scale well to high dimensional systems due to the curse-of-dimensionality for discrete problems. One common approach to handle large systems is to solve a tailored approximation to the system of interest. This paper introduces a generalizable, scalable approach to ill-behaved non-linear control problems.\\ 


Any control problem can be written as an optimization program with an objective to be optimized and with constraints on state variables. The goal is to compute a control $u(t)$ that minimizes the objective function:

\begin{equation}\label{eq:obj}
J(t,x(t),u(t)) = \int_{0}^T g(t,x(t),u(t))dt + \Psi(x(T))
\end{equation}
where $x(t)$ is the system state, which evolves according to the state equations

\begin{equation}\label{eq:dy}
x(t) = x(0) + \int_0^t f(\tau,x(\tau),u(\tau))d\tau, \quad x(0) = x_0,\ t\in[0, T]
\end{equation}
and the control $u(t) \in U$ is bounded. The objective function $J(t,x(t),u(t))$ is the total accumulated cost over the time horizon of the problem. It is composed of an operational running cost $g(t,x(t),u(t))$ and the operational terminal cost $\Psi(x(T))$. The function $f(\tau,x(\tau),u(\tau))$ is the continuous dynamics of the system which can be non-differentiable with respect to the state $(x(t))$ and the control variable $(u(t))$. The state set $X=\{x\}$ can in general be a non-convex set.\\

The Hamiltonian formulation for control problems was introduced by Pontryagin as part of his minimum principle \cite{athans2006optimal}. In this formulation the optimal control problem is given in terms of the Hamiltonian $\Ha(t,x(t),p(t),u(t)) = g(t,x(t),u(t)) + p^Tf(t,x(t),u(t))$ where $p(t)$ is the momentum (co-state). The necessary conditions for optimality are \\
\begin{equation}\label{eq:1}
\left(\frac{dx(t)}{dt}\right)^T = \frac{\partial \Ha(t,x(t),p(t),u(t))}{\partial p}
\end{equation}

\begin{equation}\label{eq:2}
\left(\frac{dp(t)}{dt}\right)^T = -\frac{\partial \Ha(t,x(t),p(t),u(t))}{\partial x}
\end{equation}

\begin{equation}\label{eq:3}
\Ha(t, x^*(t),p(t),u^*(t)) \leq \Ha(t,x(t),p(t),u(t))
\end{equation}

and optimal $u^*(t)$, and where $p(T) = \frac{\partial \Psi}{\partial x}\Big|_{x(T)}$ and $x(0) = x_0$.\\

Thus we seek an optimal trajectory $u^*(t)$ such that (\ref{eq:1}), (\ref{eq:2}) and (\ref{eq:3}) hold.\\

Solving (\ref{eq:1}), (\ref{eq:2}) and (\ref{eq:3}) directly is a two point boundary value problem because $x(0)$ and $p(T)$ are known. Unfortunately, there are no analytic solutions to solve such problems. Additionally, Pontryagin's necessary conditions of optimality are only valid for continuous systems. \\

The chattering approach approximates a solution to the optimal control problem defined by (\ref{eq:1}), (\ref{eq:2}) and (\ref{eq:3}) that are non-linear, non-convex and non-smooth. The chattering problem can also be applied to discrete systems by \textit{continualizing} the discrete system. A good description of the advantages of approximating a discrete dynamical system using a continuous system model is discussed in the article by \citeN{disc2cont}.  \citeN{kohn1999multiple} provide a formal method to find a continuous approximation to discrete systems.\\

Section~\ref{sec:ch} describes the chattering approximation approach. Section \ref{sec:alg} provides a numerical method, we call the chattering algorithm. Section~\ref{sec:results} demonstrates this approach on a real time enterprise scheduling and control system.

 \section{Chattering approach to control}\label{sec:ch}
The chattering approach to optimal control problems is motivated through measure based controls. \citeN{Kohn2010736} provide an approach for optimization of a specific class of variational problems via chattering. This paper extends the chattering lemma \cite{Kohn2010736} to a chattering approach for generic optimal control problems. The main idea in this paper is to work directly with control Hamiltonians by converting (\ref{eq:1}), (\ref{eq:2}) and (\ref{eq:3}) into a sequence of standard variational problems. Each variational problem is defined over a small time interval $\Delta_{t_i}$ at time $t_i$, $i = 0,\cdots, I-1$, by partitioning $[0, T]$ into $I$ intervals $[t_i, t_{i+1}]$ defined for each $i$ with $\Delta_{t_i} = t_{i+1} - t_i$ and such that $t_0 = 0$ and $t_I = T$.\\

The necessary conditions in (\ref{eq:3}) are relaxed by defining the optimization problem over a small time interval $\Delta {t_i}$, $\Delta {t_i} = t_{i+1}-t_i$, where the objective function is an accumulation of the Hamiltonian and the constraints are obtained by integrating (\ref{eq:2}) and (\ref{eq:3}) over the same interval $[t_i,t_{i+1}]$ for every $i$. The approach consists of finding a solution to the following sequence of variational problems.
\begin{align}\label{var}
\begin{split}
\inf_{u(t) \in U,\ t \in [t_0, t_{1}]} & \int_{t_0}^{ t_1}\Ha(t, x(t),p(t),u(t))dt\\
\text{subject to} & \hfill \\
x(t) = x_{t_0} + &\int_{t_{0}}^{t}\frac{\partial \Ha(t,x(t),p(t),u(t))}{\partial p}dt\\
p(t) = p_{t_1} - & \int^{t}_{t_1} \frac{\partial \Ha(t,x(t),p(t),u(t))}{\partial x}dt
\end{split}
\end{align}

\begin{align}\label{var:1}
\begin{split}
\inf_{u(t) \in U,\ t \in [t_1, t_2]} & \int_{t_1}^{ t_2}\Ha(t, x(t),p(t),u(t))dt \\ 
\text{subject to} & \hfill \\
x(t) = x_{t_1} + &\int_{t_{1}}^t\frac{\partial \Ha(t,x(t),p(t),u(t))}{\partial p}dt\\
p(t) = p_{t_2} - & \int^t_{t_2} \frac{\partial \Ha(t,x(t),p(t),u(t))}{\partial x}dt
\end{split}
\end{align}
$$
\vdots \\
$$
\begin{align}\label{var:2}
\begin{split}
\inf_{u(t) \in U,\ t \in [t_{I-1}, t_{I}]} & \int_{t_{I-1}}^{ t_I}\Ha(t, x(t),p(t),u(t))dt\\
\text{subject to} & \hfill\\
x(t) = x_{t_{I-1}} + &\int_{t_I}^{t}\frac{\partial \Ha(t,x(t),p(t),u(t))}{\partial p} dt\\
p(t) = p_{t_{I}} - & \int^{t}_{t_I} \frac{\partial \Ha(t,x(t),p(t),u(t))}{\partial x}dt.
\end{split}
\end{align}
with initial condition $x(0) = x_0$ and terminal transversality condition $p(T) = \frac{\partial \Psi}{\partial x}\Big|_{x(T)}$.

We apply the chattering lemma to convert the sequence of variational problems in (\ref{var}), (\ref{var:1}) and (\ref{var:2}) over the interval $[0,T]$ to a sequence of linear optimization problems over time intervals $[t_i,t_{i+1}],\forall i$.

\begin{lemma}\label{lemma}
(Chattering Lemma c.f. \cite{Kohn2010736,kohn2003control,roubicek1997relaxation}) Let $c^1({t_i}), \cdots, c^K(t_i)$ be given integrable functions from interval $[t_i, t_{i+1}]$ into $\mathbb{R}^n$ with compact support and let a constant $\epsilon > 0$ be given. Then there exists a chattering combination $g_{t_i}^\alpha$ defined for any $\alpha^{t_i} = (\alpha^{t_i}_1,\cdots,\alpha^{t_i}_K) \in \mathbb{R}^K_+$,with $\sum_{k=1}^K\alpha_k^{t_i}=1$ and $\alpha^{t_i}_k \geq 0$ such that
$$
\max_{t\in[t_i,t_{i+1}]}\left|\int_{t_i}^t\left(g_{t_i}^\alpha(\tau)-\sum_{k=1}^K\alpha_k^{t_i}c^{t_i}_k\right)d\tau\right|<\epsilon
$$
and
$$
meas\{t:g_{t_i}^\alpha(t)\neq g^\beta(t)\} \rightarrow 0,\ as,\ \beta \rightarrow \alpha_{t_i},\ \forall \beta
$$
\end{lemma}.

\begin{figure}[!htbp]
  \centering
    \includegraphics[width=0.80\textwidth]{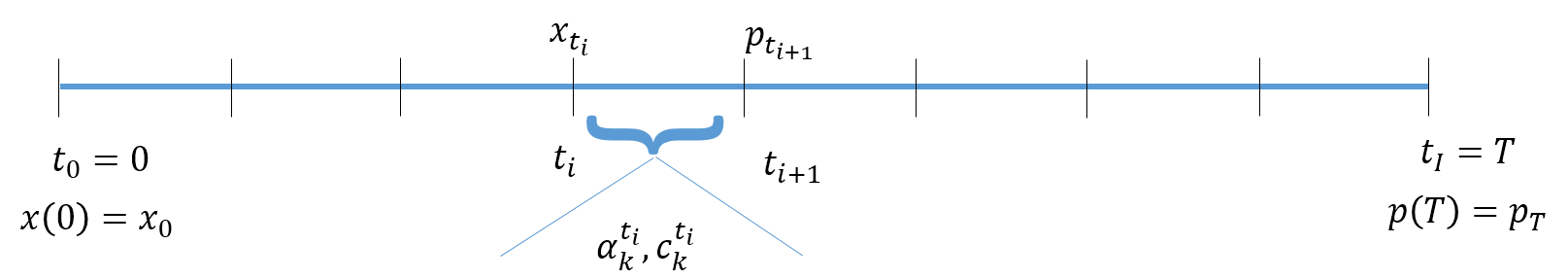}
    \caption{Partitioning Schema: $x_{t_i}$, $p_{t_i}$ are instantiated at the beginning of the interval $[t_i,t_{i+1}]$ while the chattering function $c_{t_i}^k$ and measure $\alpha_{t_i}^k$ are defined over the interval $[t_i,t_{i+1}]$.}
		\label{fig:partition}
\end{figure} 

Figure~\ref{fig:partition} provides a graphical representation of the partitioning schema, where $x({t_i}) = x_{t_i}$, $p({t_i})=p_{t_i}$ are instantiated at the beginning of the interval $[t_i,t_{i+1}]$ while the chattering function $c^{t_i}_k$ and measure $\alpha^{t_i}_k$ are defined over the interval $[t_i, t_{i+1}]$. The chattering levels $c^{t_i}_k$ can be any function over the interval $[t_i,t_{i+1}]$ but we assume them to be constant for the purpose of this paper.

\begin{figure}[!htbp]
  \centering
    \includegraphics[width=0.60\textwidth]{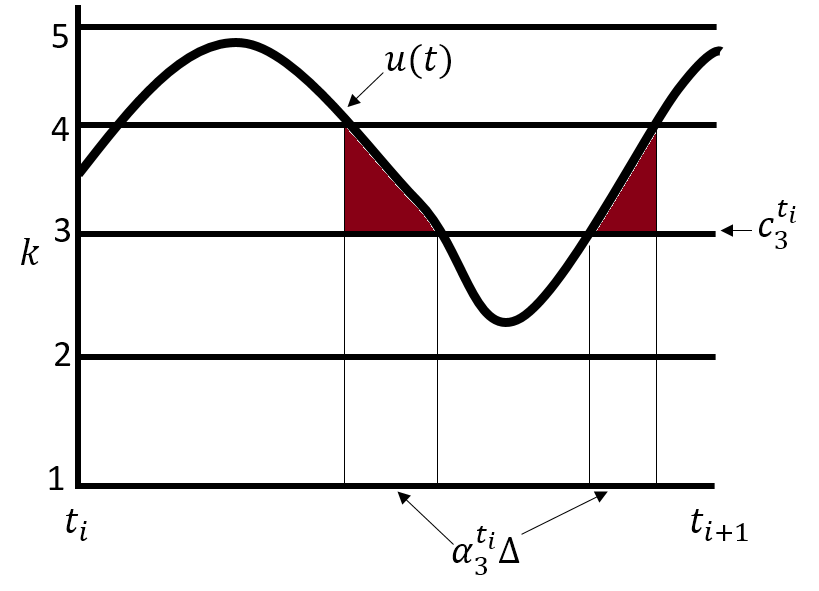}
    \caption{Graphical description of chattering}
		\label{fig:chattering}
\end{figure}

Figure~\ref{fig:chattering} gives a graphical representation of the chattering in the interval $[t_i,t_{i+1}]$. The chattering measure $\alpha^{t_i}_k=\alpha^{t_i}(s)$, where $s = [k,k+1]$ for some $k$, can be interpreted as a probability that in an infinitesimally small neighborhood $\Delta_{t}$ around $t$, we can find $c_k^{t_i} (\tau) \in s$ for all $\tau \in \Delta_{t}$. For illustration $k=1,\cdots,5$, $s=[3,4]$, $t=t_i$ and $\Delta_t = t_{i+1} - t_i$ in Figure~\ref{fig:chattering}.\\

The following theorem enables us to derive the chattering formulation for the optimization problem in (\ref{var}), (\ref{var:1}) and (\ref{var:2}).

\begin{theorem}\label{th1}
Consider the sequence of variational problems in (\ref{var}), (\ref{var:1}) and (\ref{var:2}), $i=\{1,\cdots, I-1\}$. For each $[t_i, t_{i+1}]$, define levels $c_k^{t_i}, k = 1, \cdots, K$ that discretize the range of $u \in U$ such that they form an ordered bounded set. Then there exist $\alpha_k^{t_i}, k = 1, \cdots, K; i = 0, \cdots, I-1$ and $\sum_{k=1}^K \alpha_k^{t_i} = 1, \forall i$ such that the chattering combination $g_{t_i}^\alpha$  for $t \in \Delta_k^{t_i}$ closely approximates, in a Lebesgue norm sense, a solution $u^*(t)$ to the variational problem, and, moreover, 
$$
\int_{t_i}^{ t_i+\Delta_{t_i}}\Ha(t, x(t),p(t),u(t))dt \approx  \sum_{k=1}^K\Ha(t_i, x_{t_i},p_{t_i},c_k^{t_i})\alpha_k^{t_i}\Delta_{t_i}.
$$
where $t_{i+1} = t_i + \Delta_{t_i}$ and $x(t_i) = x_{t_i}$ and $p(t_i) = p_{t_i}$
\end{theorem}

{\bf Proof}. Let us consider a sequence $c_k^{t_i}, k = 1, \cdots, K$ in $U$. Define a Young's measure $\alpha(t,S)$, for $S \subseteq U$, as the limit of the sequence as $k \rightarrow \infty$. We associate with $\alpha(t_i, s), \forall s \in S$ a constant measure $\alpha^{t_i}(s)$, such that $\alpha(t_i, s) = \alpha^{t_i}(s)$ for $t \in [t_i,t_{i+1}]$ and $\alpha^{t_i}(s) = \sum_{k=1}^K \alpha_k^{t_i}\delta(c_k^{t_i})$, where the levels $c_k^{t_i}, k = 1, \cdots, K$ discretize the range of $u \in U, \sum_{k=1}^K \alpha_k^{t_i} = 1$ and $\alpha^{t_i}_k \geq 0$, $\forall i$ and $\delta(c_k^{t_i})$ are Dirac measures concentrated on $c_k^{t_i}$. Then

$$
u(t) = \int_U s\alpha^{t_k}(ds)
$$
where the integral is written as (with a slight abuse of notation)

$$
\int_U s\sum_{k=1}^K\alpha^{t_i}(s)\delta(c_k^{t_i}) = \int_U \alpha^{t_i}(s)\sum_{k=1}^Ks\delta(c_k^{t_i}) = \sum_{k=1}\alpha_k^{t_i}c_k^{t_i}.
$$

This implies that the last sum gives a close approximation to the solution of the relaxed problem in (\ref{var}), (\ref{var:1}) and (\ref{var:2}) which is a close approximation to the solution of the original variational problem (\ref{eq:obj}) and (\ref{eq:dy}). Applying the chattering lemma the last sum above is arbitrarily closely approximated by the chattering combination for all $i$.\\

Therefore, by the properties of a Lebesgue integral and $\Ha$ continuous in $(x,p)$ and piecewise continuous in $u$, for any $u \in U$,
$$
\int_{t_i}^{ t_i+\Delta_{t_i}}\Ha(t, x(t),p(t),u(t))dt \approx \sum_k \Ha(t_i, x_{t_i},p_{t_i},c_k^{t_i})\alpha_k^{t_i}\Delta_{t_i}
$$
for small enough $\Delta_{t_i}$.$\qed$\\

The chattering approximation of $u(t)$ on $t$ yields

\begin{equation}\label{eq:propu}
\int_{t_i}^{t_{i+1}}u(s)ds \approx \sum_{k=1}\alpha_k^{t_i}c_k^{t_i}
\end{equation}
for $i = 0, \cdots, I-1$, which also implies that the chattering approximation to equations (\ref{eq:1}) and (\ref{eq:2}) (in the integral form) can be written as

\begin{equation}\label{eq:propx}
x_{t_{i+1}} = x_{t_i} + \sum_{k=1}^K\alpha_k^{t_i} \frac{\partial \Ha(t,x(t),p(t),c_k^{t})}{\partial p}\Big|_{t=t_i}\Delta_{t_i}
\end{equation}

\begin{equation}\label{eq:propp}
p_{t_{i+1}} = p_{t_i} - \sum_{k=1}^K\alpha_k^{t_i} \frac{\partial \Ha(t,x(t),p(t),c_k^{t})}{\partial x}\Big|_{t=t_i}\Delta_{t_i}
\end{equation}

Note that this chattering approximation can in general be written for any variable whose evolution depends on the variable $u(t)$.

\section{Linear optimization problem and numerical method}\label{sec:alg}
Now we use the chattering approximation in Section~\ref{sec:ch} to formulate a sequence of linear optimization programs where we optimize over $\alpha$ for each time interval  indexed by $i$.\\ 

The linear optimization program based on the approximations in Theorem~\ref{th1} is

\begin{equation}\label{opt:1}
\min_{\alpha_k^{t_i}} \sum_k\Ha(t_i,x_{t_i},p_{t_i},c_k^{t_i})\alpha_k^{t_i}
\end{equation}

subject to

\begin{equation}\label{opt:2}
\sum_k \alpha_k^{t_i} = 1
\end{equation}

\begin{equation}\label{opt:3}
0 \leq \alpha_k^{t_i} \leq 1,\ \forall k
\end{equation}
given $x_{t_i}, p_{t_i}$ and known $c_k^{t_i}$, for each $i = 0, \cdots I-1$. This is a relaxed knapsack problem which has an analytic solution. Hence, there is no computational cost in solving this linear program.\\

The control problem can now be solved using the above chattering formulation and the known initial condition on the state ($x(0)=x_0$) and a good estimate of the initial value of the co-state. In practical problems it is almost impossible to know apriori a good estimate of the initial co-state. In order to find a good estimate of the initial condition we use a similar procedure as in the numerical method of \textit{variation of extremals} (see \cite{kirk2004optimal}) but while solving the chattering optimization program described above. This approach converts the two point boundary value problem into an initial value problem, which is much easier to solve numerically.\\

The algorithm starts by guessing the initial value of the co-state ($p(0)=p_0$) and iteratively corrects on this guess based on the final observation $p(T)$. This adaptation is done by considering a perturbation on the initial condition of the co-state. Given $(x_{t_i},p_{t_i})$ and the chattering levels $c^k_{t_i}$, the optimization program given by (\ref{opt:1}), (\ref{opt:2}) and (\ref{opt:3}) is solved to obtain the optimal chattering measure $\alpha_{t_i}^k$. The chattering levels and the chattering measure along with equations~(\ref{eq:propx}) and (\ref{eq:propp}) are used to obtain the trajectory $(x_{t_i},p_{t_i})$.\\

In order to propagate $(x_{t_i},p_{t_i})$ starting at $(x_0,p_0)$, the algorithm defines the chattering levels $c_k^{t_i}$ for each time interval from index $i$ to $i+1$, such that they cover the control set $U$ and that any bounded constraints on $x_{t_{i+1}}, \forall i$ are not violated. Additionally, in order to find  a suitable bound for the levels, $c_{max}^i = \max_k c_k^{t_i},\ \forall i$ and $c_{min} = \min_k c_k^{t_i}$, the following equation has to be satisfied

\begin{equation}\label{eq:bound}
x_{min} \leq x_{t_i} + f(x_{t_i},u_{t_i})\Delta_{t_i} \leq x_{max}.
\end{equation}

Given $x_{min}$ and $x_{max}$, bounds on chattering levels can be determined nuemrically from (\ref{eq:bound}). This will ensure that the chattering levels cover the set $U$ and the state constraints are satisfied.\\

The algorithm also propagates a perturbed trajectory that starts at the initial state $x_{t_i}^\delta = x_0$ and the perturbed initial co-state $p^\delta_0 = p_0 + \delta p$, for small $\delta p$. The corresponding trajectory for state and co-state, starting at $(x_0,p_0^\delta)$ at index $i$   is given by $(x_{t_i}^\delta, p_{t_i}^\delta)$.

The algorithm then corrects the initial estimate of co-state by adapting the guesstimate based on deviation of the propagated co-state ($p(T)$) starting from $p_0$ from the actual final value of the co-state $\frac{\partial \Psi}{\partial x}\Big|_{x(T)}$. This correction is done using 
$$
p_0 \leftarrow p_0 + \gamma \left(\left[\frac{\partial^2 \Psi}{\partial x^2}({x(T)})\right]P_x(T) - P_p(T)\right)^{-1}\left(p_T - \frac{\partial \Psi}{\partial x}({x(T))}\right)
$$ 
where $\gamma$ is a step parameter, $P_x(t_i) = \frac{\partial x}{\partial p_0}\Big|_{x = x_{t_i},p=p_{t_i}}$ and $P_p(t_i) = \frac{\partial p}{\partial p_0}\Big|_{x = x_{t_i},p=p_{t_i}}$. The partial derivative can be numerically estimated by using a finite difference method. This process is repeated until convergence, i.e., $\norm{p_T - \frac{\partial \Psi}{\partial x}({x(T)})} < \epsilon$. At convergence the solution obtained for $x_{t_i}$ is the optimal control trajectory and the solution obtained for $u_{t_i} = \sum_{k=1}\alpha_k^{t_i}c_k^{t_i}$ is the optimal control signal. The triple $\{t_i,x_{t_i},u_{t_i}\}$ defines a control law (policy).\\

The procedure described above gives a simple routine for solving the optimal control problem, which is given by Algorithm~\ref{alg}.

\begin{algorithm}[!htbp]
\SetAlgoLined
\KwIn{Historical data containing the sequence $\{t,x(t),u(t),x'(t)\}$, where $x'(t)$ is the next state obtained on applying control $u(t)$ at time $t$ while in state $x(t)$. The criterion ($g(t,x(t),u(t)$ and $\Psi(x)$) that needs to be optimized.}
\KwResult{A sequence of (near) optimal control law $\{t_i,x_{t_i},u_{t_i}\}$ for $i = 0,\cdots,I$. }
\Begin{
Estimate the Hamiltonian $\Ha(t,x(t),p(t),u(t))$ from the given data or estimate the dynamics $f$ (using generalized Kalman filter or any other equivalent method) and compute $\Ha = g(t,x(t),u(t)) + p^Tf(t,x,u)$\;
If the system is discrete use a continuous approximation (via \emph{continualization})\;
Initialize $i=0,\ t_i=0,\ x(0)=x_0$ and guess $p(0)=p_0$ and $p_0^\delta = p_0 + \delta p$ for a fixed $\delta p$\;
flag = False.\;
\While{flag is False or $iter \leq maximumIterations$}{
\While{$i < I$}{
	\emph{Measure $x_{t_i}$}\;
	\emph{Determine chattering levels $c_k^{t_i}$ such that they satisfy bounded constraints on $u(t)$}\;
	\emph{Solve the optimization program (\ref{opt:1}), (\ref{opt:2}) and (\ref{opt:3}) twice, once with $(x_{t_i}, p_{t_i})$ and second one with $(x_{t_i}^\delta, p_{t_i}^\delta)$}\;
	 \emph{Compute $u_{t_i} = \sum_{k=1}^K\alpha_k^{t_i}c_k^{t_i}$}\;
	 \emph{Compute $p_{t_{i+1}},\ p_{t_{i+1}}^\delta,\ x_{t_{i+1}}^\delta$ using (\ref{eq:propx}) and (\ref{eq:propp})}\;
	 \emph{Store $x_{t_i},\ p_{t_i},\ u_{t_i}$}
	 }
	 \eIf{$\norm{p_T - \frac{\partial \Psi}{\partial x}({x(T)})} < \epsilon$}{flag = True}{
	  \emph{Compute $P_x(T) = \frac{x_T^\delta - x_T}{\delta p},\ P_p(T) = \frac{p_T^\delta - p_T}{\delta p}$}\;
	 $p_0 \leftarrow p_0 + \gamma \left(\left[\frac{\partial^2 \Psi}{\partial x^2}({x(T)})\right]P_x(T) - P_p(T)\right)^{-1}\left(p_T - \frac{\partial \Psi}{\partial x}({x(T))}\right)$}
	 }
	 }
	 \caption{Algorithm for Hamiltonian-Control via Chattering}
	 \label{alg}
\end{algorithm}

\section{Results}\label{sec:results}
This section presents two  applications that demonstrate the chattering control method described in this paper.  The first example is used as a proof-of-concept. We use a standard linear quadratic oscillator to compare the solution from the chattering algorithm (Alg. \ref{alg}) described in this paper with the known analytic solution. The second demonstration is on a real-time enterprise scheduling and control process. This enterprise problem was first introduced by \citeN{doi:10.1080/07408170590516953}. This problem is a hybrid of continuous and discrete systems that is highly non-linear, non-convex and non-smooth.

\subsection{Linear Quadratic Controller}
A linear quadratic controller is a system with quadratic operational cost that is constrained to move on a linear manifold.\\

\begin{figure}[!htbp]
\centering     
\subfigure[State Trajectory]{\label{fig:lq}\includegraphics[width=0.90\textwidth]{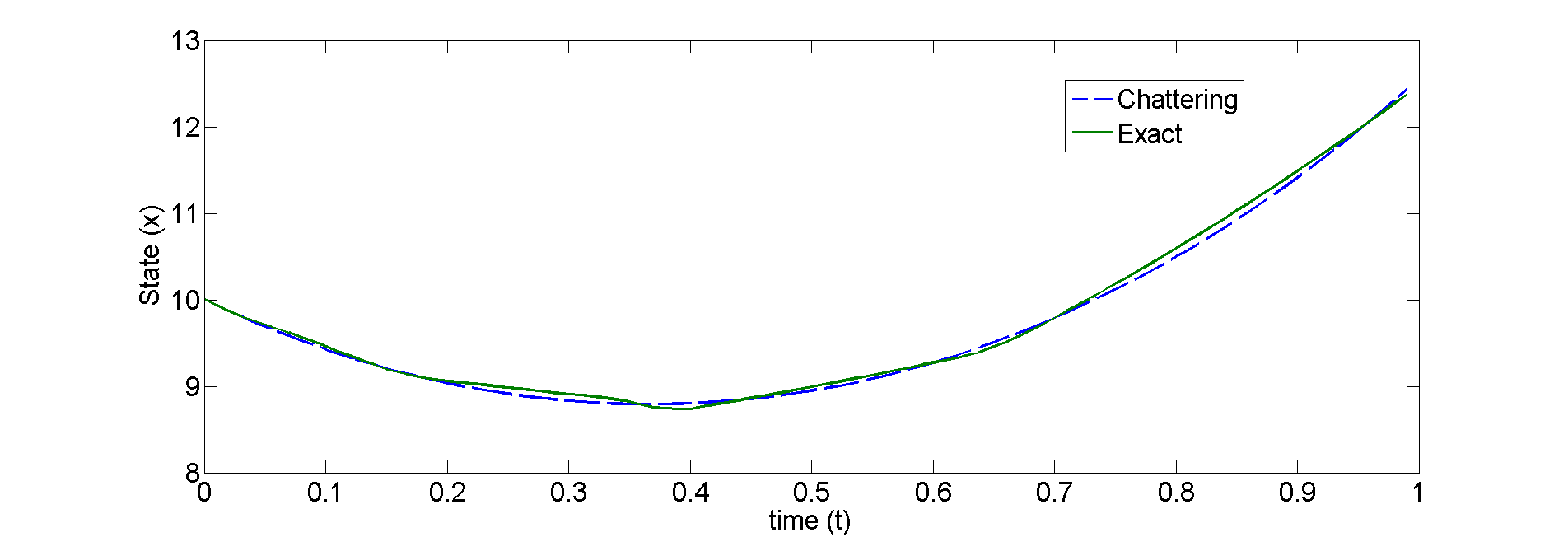}}
\subfigure[Control]{\label{fig:lq_control}\includegraphics[width=0.90\textwidth]{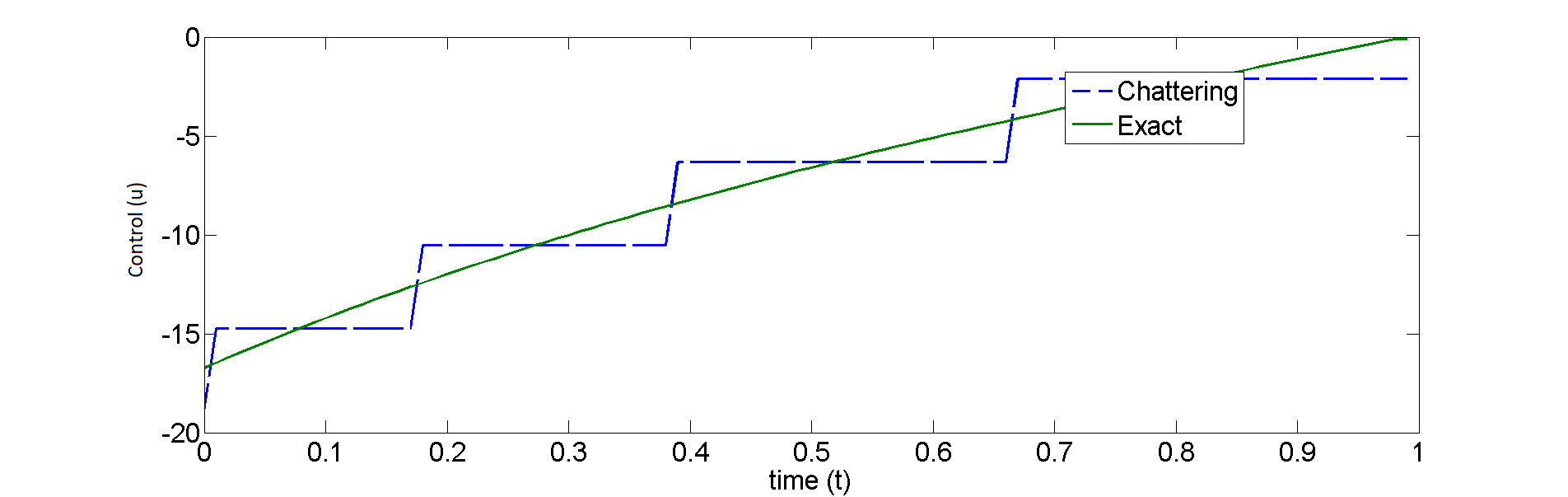}}
\caption{Linear Quadratic Controller}
\label{fig:quad}
\end{figure}

The control problem is
$$
\min_{u(t)}\int_0^1 x(t)^2 + u(t)^2 dt\
$$

s.t.
$$
\dot{x}(t) = x(t) + u(t)
$$
with initial condition $x(0) = 10$.\\

The Hamiltonian for the linear quadratic controller is given by
$$
\Ha(x,p,u) = x^2 + u^2 + p(x+u)
$$
Figure~\ref{fig:quad} compares the state trajectory and the near-optimal control obtained using the chattering method vs the state trajectory and optimal control obtained through the analytic solution. \\

As observed the chattering method provides a good approximation to the optimal state trajectory, see Fig. \ref{fig:lq}. The \textit{chattering} artifact can be observed in Fig. \ref{fig:lq_control}, where the controller chatters on discretized control levels.

\subsection{Real Time Enterprise Scheduling and Control Problem}
\begin{figure}[!htbp]
  \centering
    \includegraphics[width=0.80\textwidth]{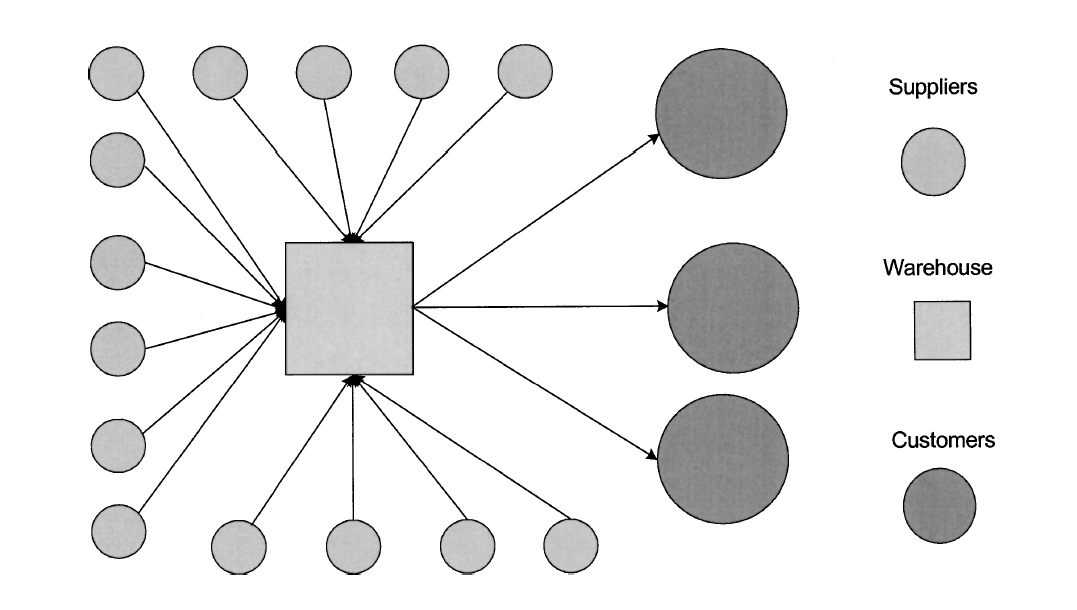}
    \caption{A distribution system including suppliers, warehouses, and customers.}
		\label{fig:grocer}
\end{figure}

The second example uses a modified version of the problem defined by \citeN{doi:10.1080/07408170590516953}. Consider a food distribution system with 14 suppliers, one warehouse, and three customers (graphically illustrated in Figure~\ref{fig:grocer}).

\begin{figure}[!htbp]
\centering     
\subfigure[Apples]{\label{fig:demand_apples}\includegraphics[width=0.49\textwidth]{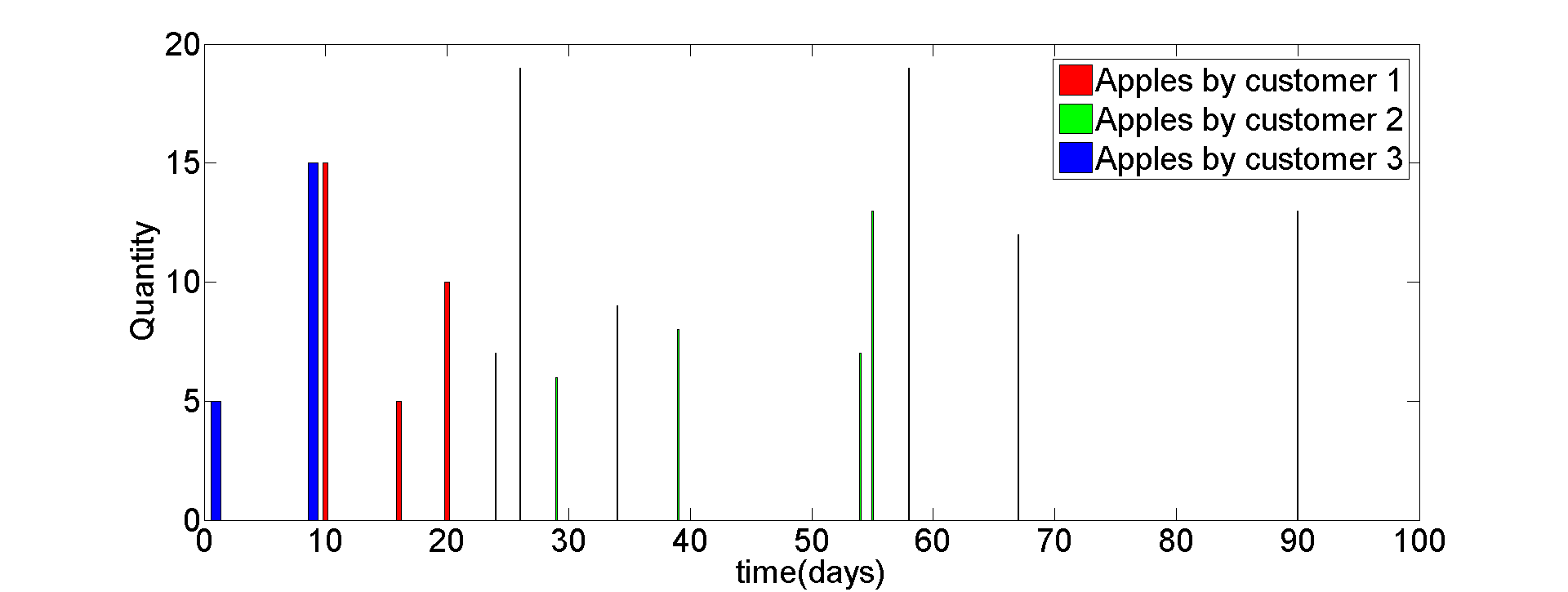}}
\subfigure[Oranges]{\label{fig:demand_oranges}\includegraphics[width=0.49\textwidth]{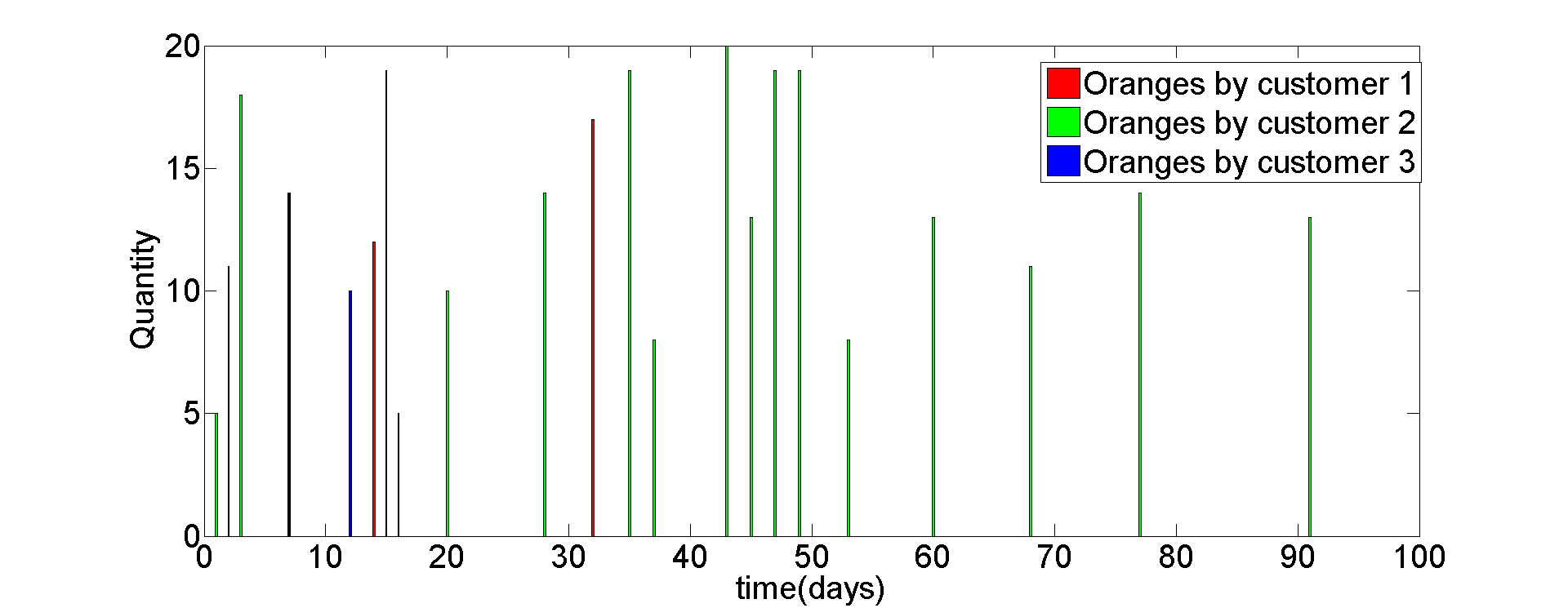}}
\subfigure[Banana]{\label{fig:demand_banana}\includegraphics[width=0.49\textwidth]{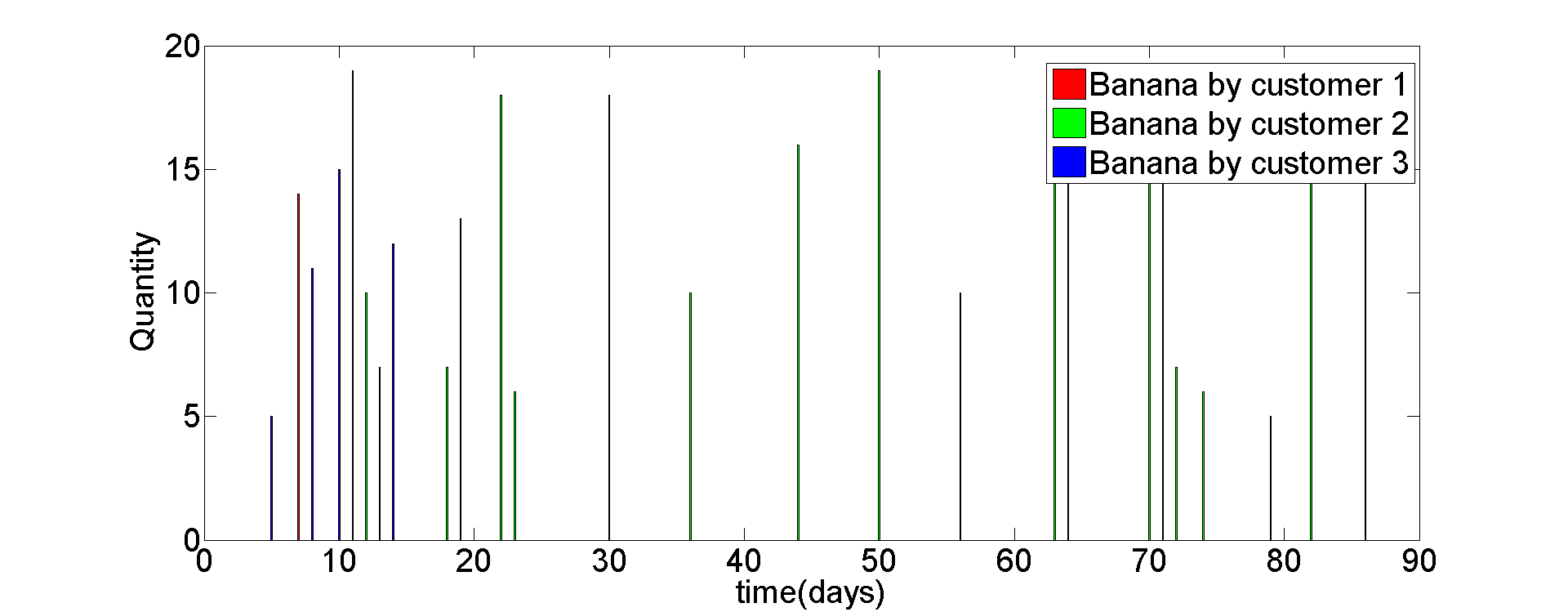}}
\subfigure[Tea]{\label{fig:demand_tea}\includegraphics[width=0.49\textwidth]{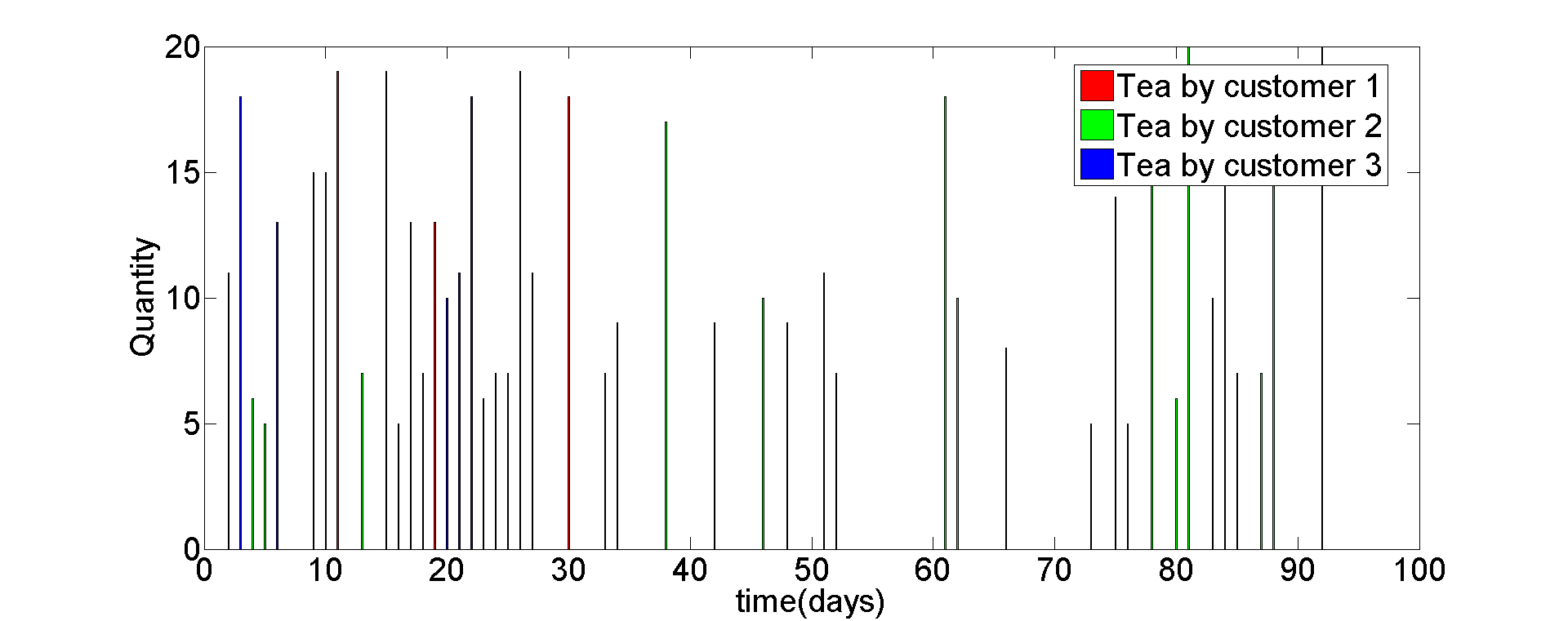}}
\subfigure[Olives]{\label{fig:demand_olives}\includegraphics[width=0.49\textwidth]{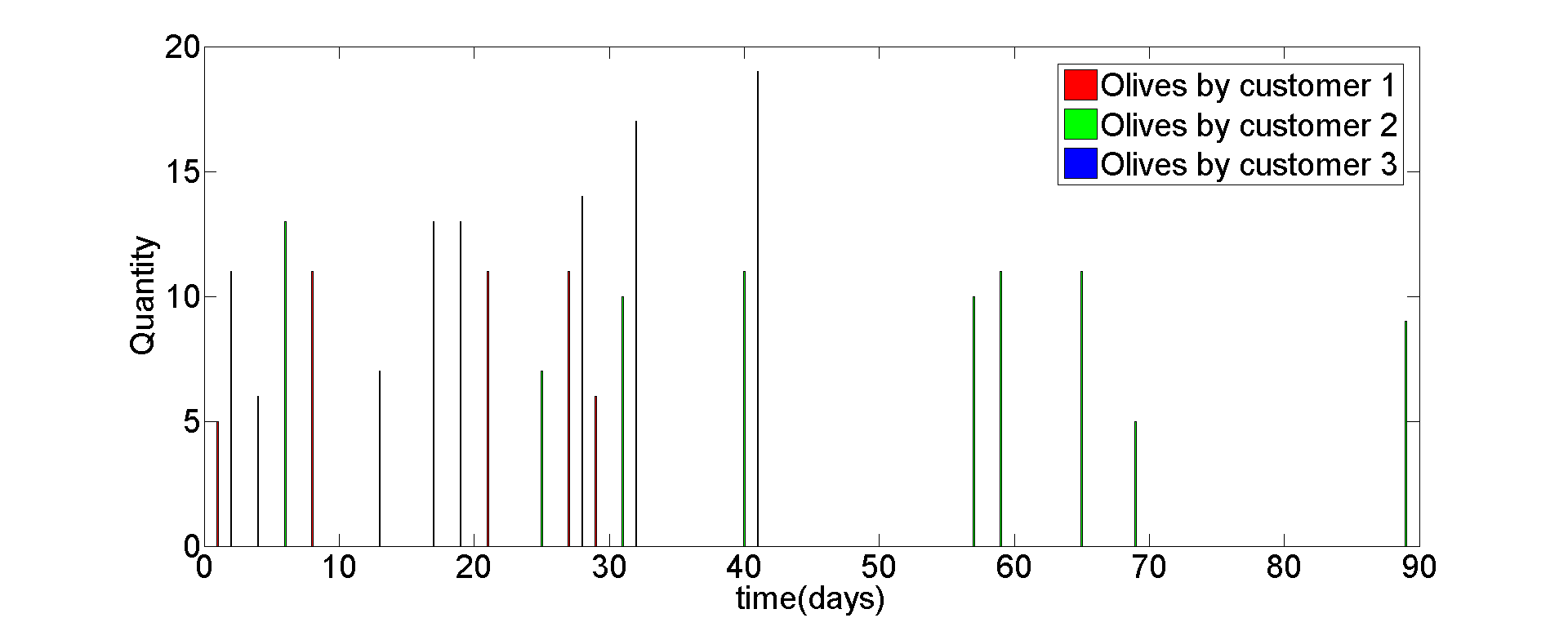}}
\caption{Demand}
\label{fig:demand}
\end{figure}

 There are five types of food items to be distributed, each type is supplied by one or more suppliers and all items are perishable. The demands by the three customers are given in Figure~\ref{fig:demand}. The customers are given an importance factor of: (i) 1.0 for most preferable, whose orders always need to be satisfied;  ii) 0.40 for average customers whose orders need to be satisfied completely in $40\%$ of the cases; and (iii) 0.25 for least preferable customers whose orders need to be satisfied completely in $25\%$ of the cases. The problem data includes: (i) a list of customers with ranking and demands for each item type; (ii) a list of suppliers with the lead times, fixed and variable costs of delivery and min and max order quantities; and (iii) a list of item types with warehouse carry-over cost per item type, penalties per unit time for not satisfying demands, and days of life before perishing. Tables \ref{tab:1} and \ref{tab:2} provide the values of these external parameters of the grocer.\\
\begin{table}
\tbl{Grocer's External Parameters\label{tab:1}}{
\begin{tabular}{|llrr|}
\hline
Item $j$ & Name & InvCarrCost $\gamma^j$ & Penalty $\delta^j$ \\
\hline
0    & Apple       & 100           & 10 \\
1    & Orange      & 150           & 25 \\
2    & Banana      & 200           & 39 \\
3    & Tea         & 50            & 30 \\
4    & Olive       & 65            & 25 \\
\hline
\end{tabular}

\quad

\begin{tabular}{|cc|}
\hline
Customer & Importance\\
$i$ & $w_i$ \\
\hline
1    & 1.0 \\
2    & 0.4 \\
3    & 0.25 \\
\hline
\end{tabular}
}
\vspace{2mm}

\tbl{Suppliers' External Parameters \label{tab:2}}{

\begin{tabular}{|llrrrr|}
\hline
Item &  Supplier &  UnitCost & FixCost & MinQty & MaxQty \\
$j$ &  $k$  & $\alpha_k^j$ & $\beta_k^j$ & $\mu_{k\ min}^j$ & $\mu_{k\ max}^j$ \\
\hline
0    &  New Hampshire        & 20     & 10     & 7      & 14 \\
0    &  Colorado             & 25     & 7      & 4      & 11 \\
1    &  Florida             & 50     & 10     & 5      & 20 \\
1    &  California           & 70     & 5      & 4      & 13 \\
2    &  Costa Rica           & 20     & 15     & 8      & 13 \\
2    &  Italy               & 30     & 20     & 2      & 13 \\
2    &  India               & 15     & 25     & 6      & 25 \\
3    &  India               & 12     & 25     & 2      & 16 \\
3    &  Sri Lanka           & 11     & 25     & 9      & 26 \\
3    &  England              & 20     & 15     & 6      & 14 \\
3    &  Market               & 23     & 20     & 2      & 18 \\
4    &  Greece              & 20     & 15     & 15     & 17 \\
4    &  Italy                & 25     & 12     & 10     & 22 \\
4    &  Market               & 30     & 18     & 11     & 14 \\
\hline
\end{tabular}

}
\end{table}

The problem is to allocate the requested quantities $v_i^j(t)$ of items ($j$) to customers ($i$) by ordering $\mu^j_k(t)$ of items $j$ from appropriate supplier $k$ while minimizing the total incremental cost. The dynamics model consists of two elements (i) market conservation; and (ii) inventory conservation. 

\begin{itemize}
\item The market conservation dynamics are given by
$$
\dot{Z}_i^j(t) = -Z_i^j(t) + \Theta_i^j(t) -v_i^j(t),\ i=1,\cdots,3,\ j=1,\cdots,5 
$$
where, $Z_i^j(t)$ is the unsatisfied demand of product $j$ to customer $i$ at time $t$, $\Theta_i^j(t)$ is the continulized demand of  product $j$ to customer $i$ at time $t$ and $v_i^j(t)$ is the continualized delivery action of product $j$ to customer $i$ at time $t$.
\item The inventory conservation dynamics are given by
$$
\dot{X}^j(t) = -X^j(t) + \sum_{k\in S_j}s_k^j(t) + \hat{\mu}^j(t) - p^j(t),\ j=1,\cdots,5 
$$
where, $X^j(t)$ is the inventory of product $j$ at time $t$, $\sum_{k\in S_j}s_k^j(t)$ is the total rate of supply of product $j$ from suppliers $k$ at time $t$, $\hat{\mu}^j(t) = \sum_{k\in S_j} \mu^j_k(t)$ is the cumulative ordering action of product $j$ from supplier $k \in S_j$ at time $t$ and $p^j(t)$ is the cumulative rate of delivery of product $j$ to customers $i$.   
\end{itemize}

The cost function to be minimized is given by

$$
J(t) = -\sum_{i=1}^{3}\sum_{j=1}^5v_i^j(t)r_i^j(t) + \sum_{j=1}^5(\alpha^j\hat{\mu}^j(t)+\beta^j) + \sum_{j=1}^5\gamma^jX^j(t) \sum_{i=1}^{3}\sum_{j=1}^5\bar{w}_i^jZ_i^j(t)
$$ 
where, $r_i^j(t)$ is the unit revenue obtained by selling product $j$ to customer $i$ at time $t$, $\gamma^j$ is the inventory cost of carrying product $j$ per unit time, $\alpha^j = \sum_k\alpha_k^j$ is the envelop unit cost of ordering product $j$, $\beta^j = \sum_k\beta_k^j$ is the envelop fixed cost of ordering product $j$, $\bar{w}_i^j = \frac{w_i\delta^j}{\sum_{i=1}^{3}\sum_{j=1}^5w_i\delta^j}$ is the cost for not satisfying demand, $\delta^j$ is the penalty for not satisfying demand for product $j$ and $w_i$ is the relative weight for not satisfying the demand for product $j$. We use the  function $J(t)\left|{J(t)}\right|$ as the objective function to be minimized (the reason we chose this function is because it provides stable and bounded solutions). For this system the state variable is $x(t) = [X(t), Z(t)]^T$. The control variable is $u(t) = [\mu(t),v(t)]^T$. The state space is continuous with dimension $d_X = 20$ and the control space is also continuous with dimension $d_U = 29$. \\

\begin{figure}[!htbp]
\centering     
\subfigure[Apples Delivery]{\label{fig:delivery_apples}\includegraphics[width=0.49\textwidth]{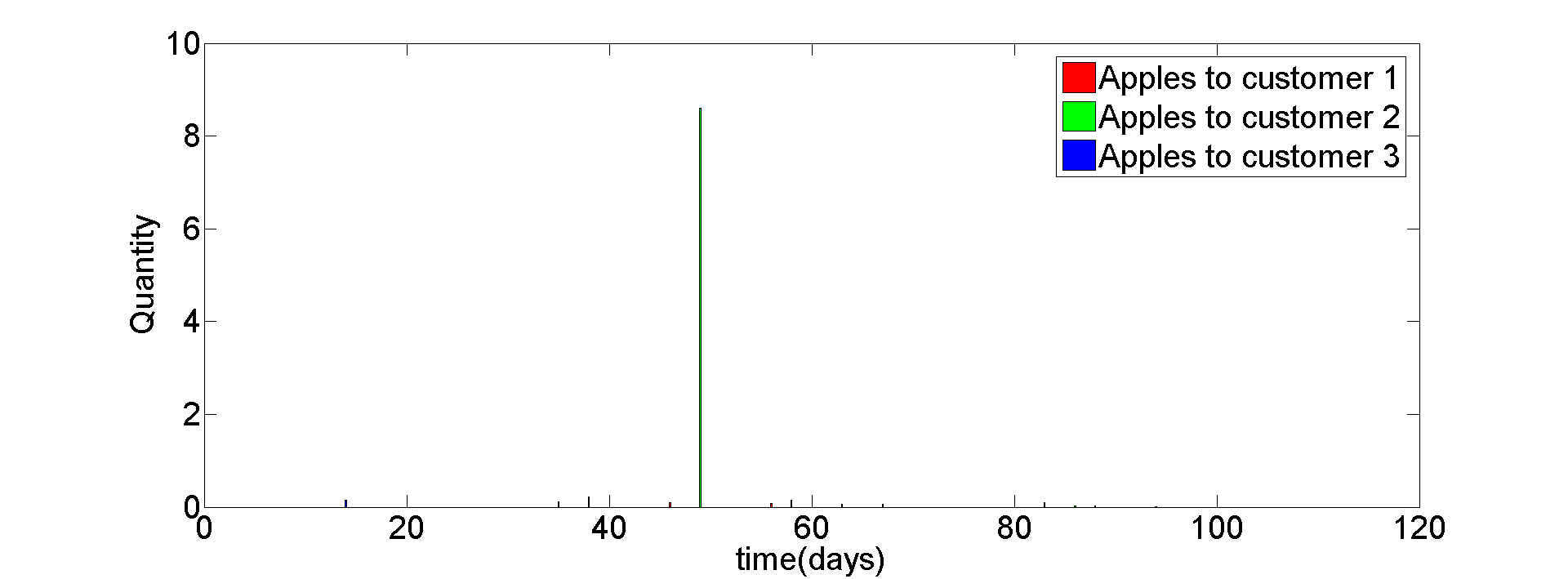}}
\subfigure[Apples Inventory]{\label{fig:inventory_apples}\includegraphics[width=0.49\textwidth]{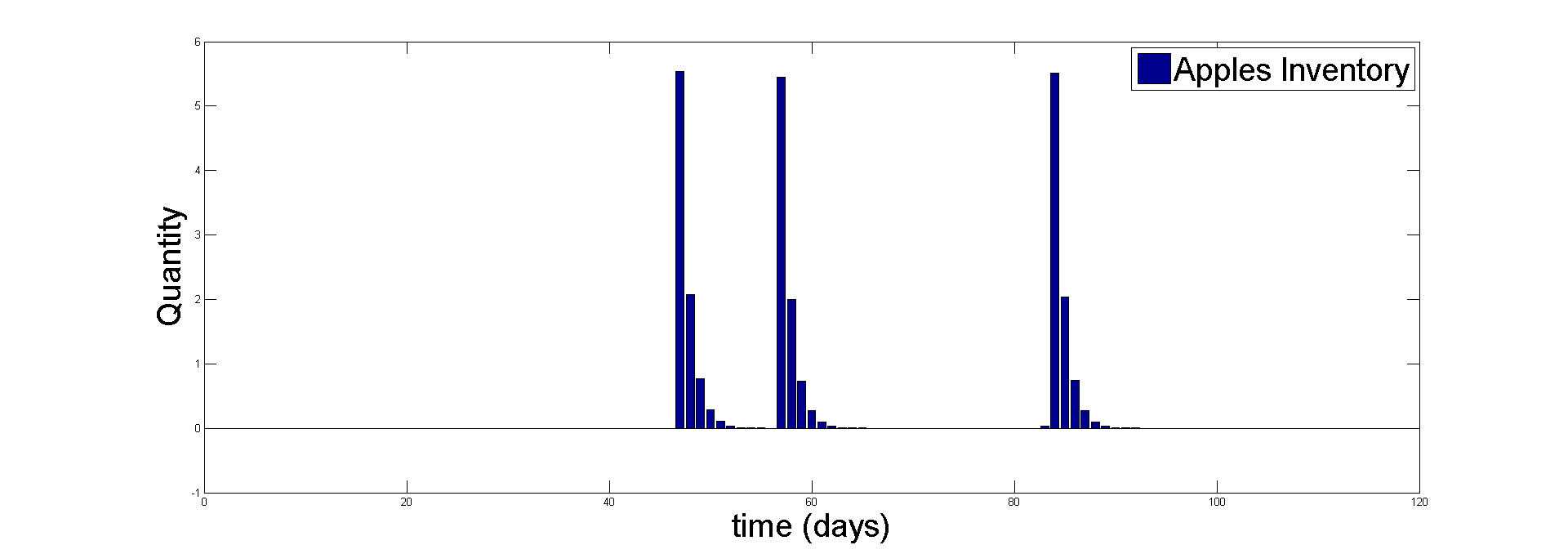}}
\subfigure[Oranges Delivery]{\label{fig:delivery_oranges}\includegraphics[width=0.49\textwidth]{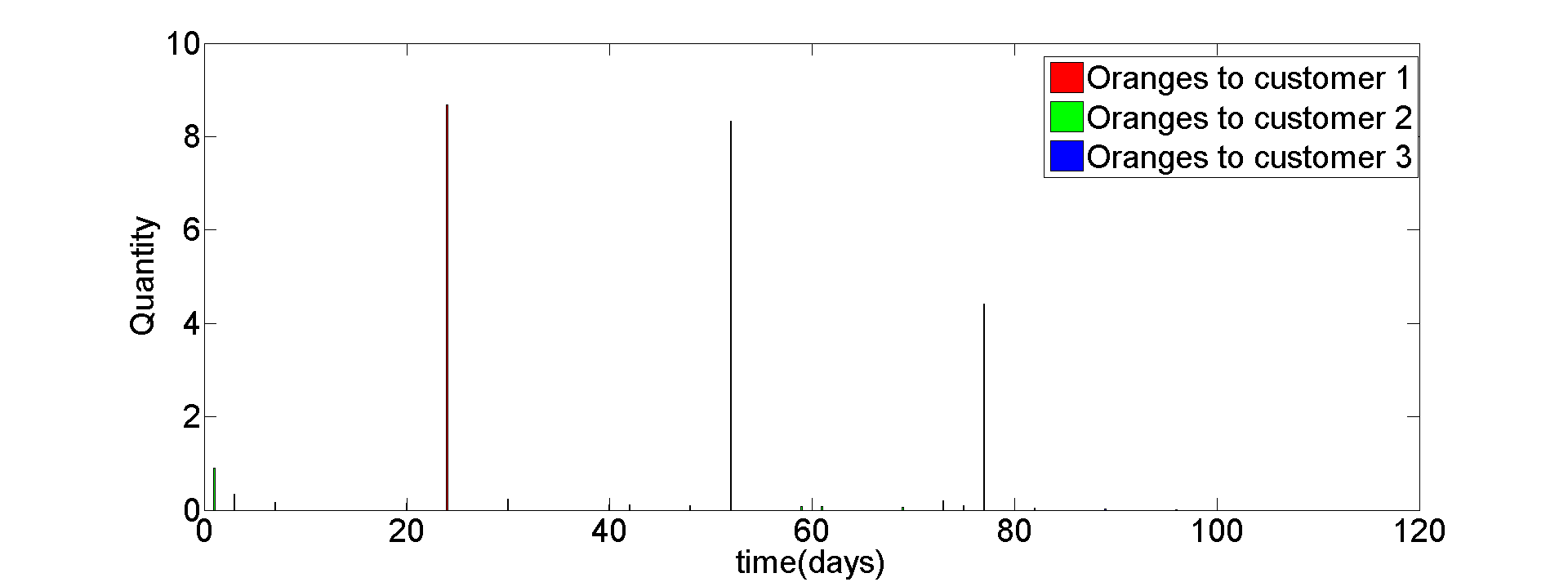}}
\subfigure[Oranges Inventory]{\label{fig:inventory_Oranges}\includegraphics[width=0.49\textwidth]{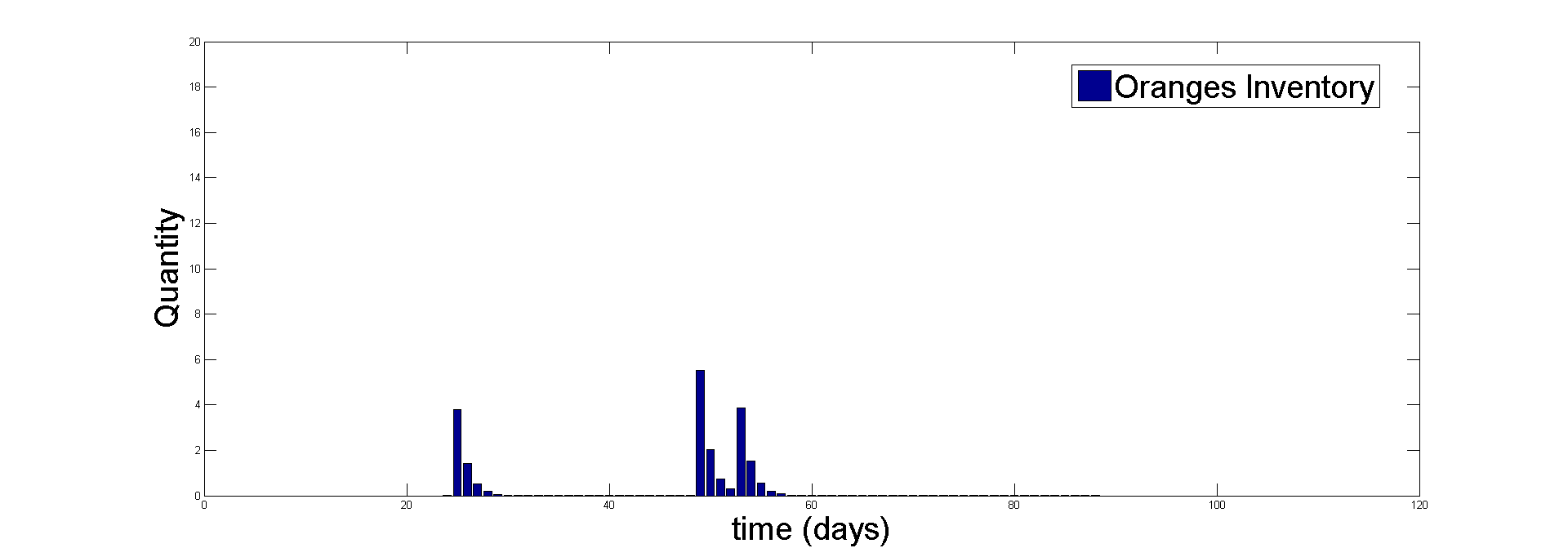}}
\subfigure[Banana Delivery]{\label{fig:delivery_Banana}\includegraphics[width=0.49\textwidth]{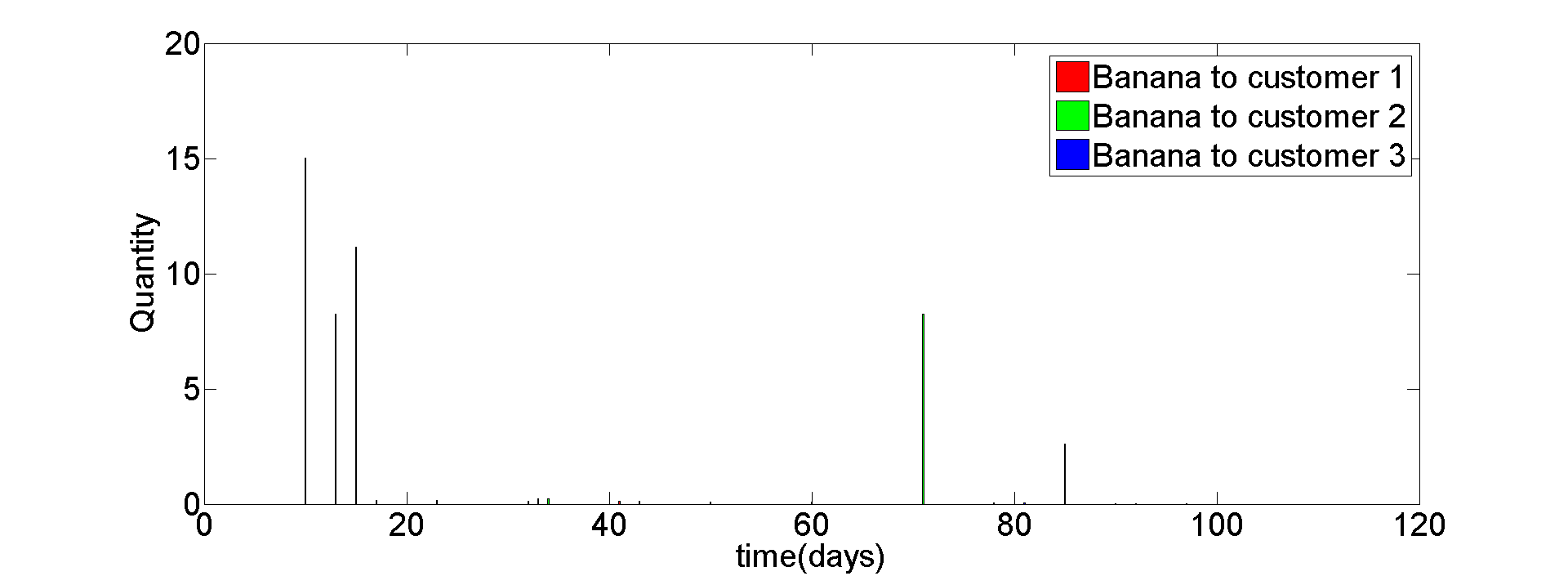}}
\subfigure[Banana Inventory]{\label{fig:inventory_Banana}\includegraphics[width=0.49\textwidth]{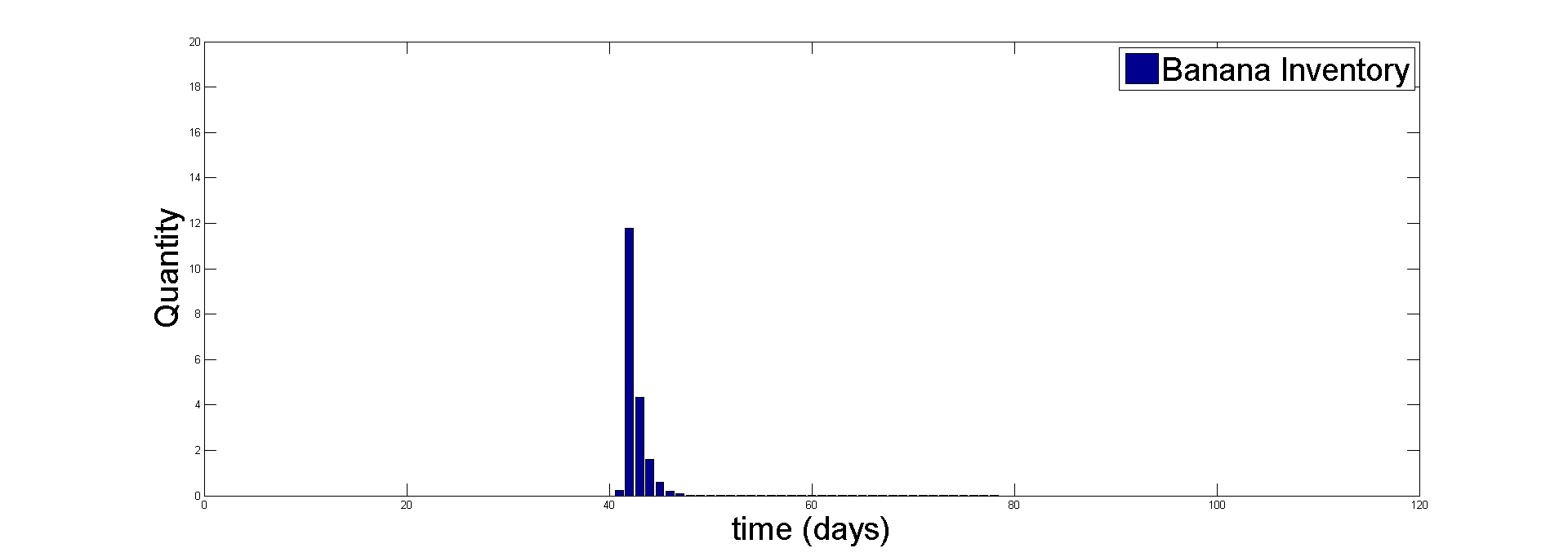}}
\subfigure[Tea Delivery]{\label{fig:delivery_Tea}\includegraphics[width=0.49\textwidth]{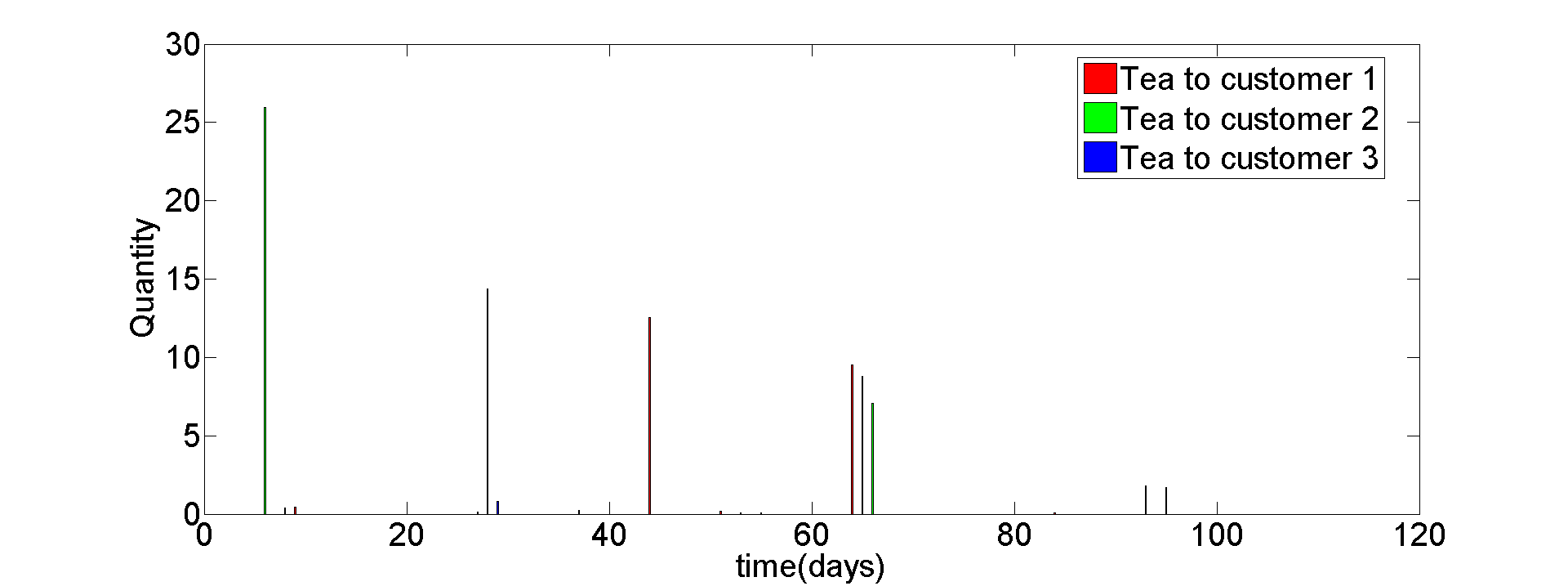}}
\subfigure[Tea Inventory]{\label{fig:inventory_Tea}\includegraphics[width=0.49\textwidth]{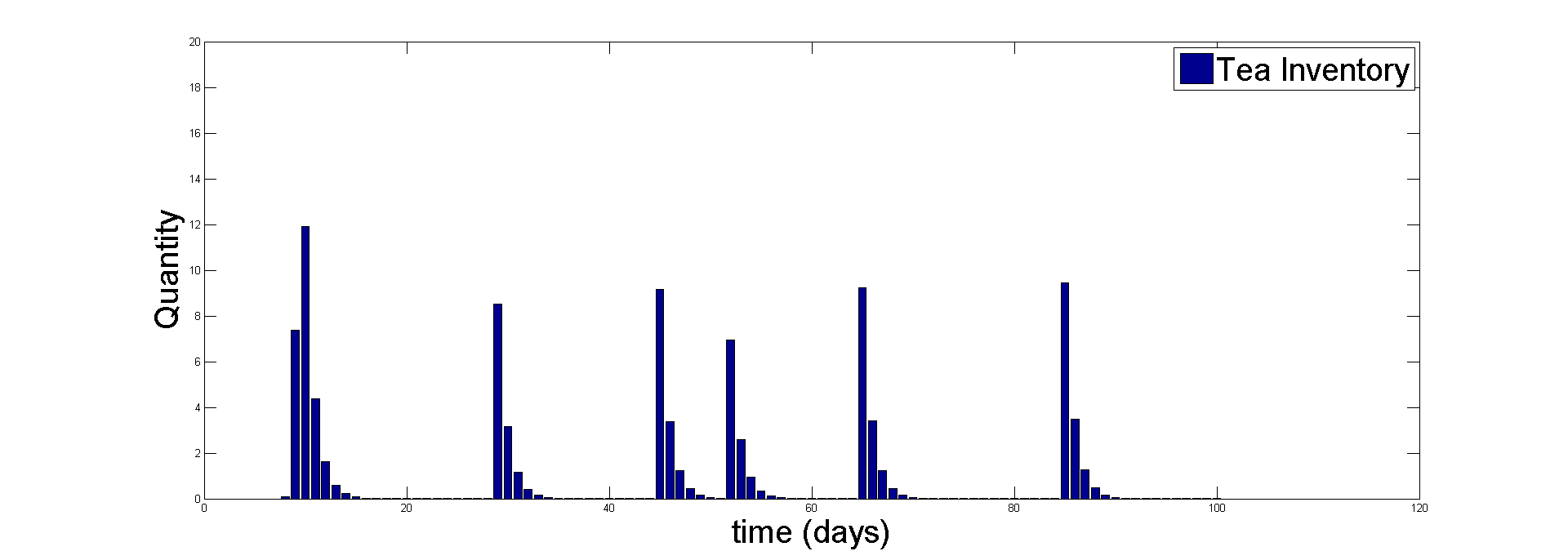}}
\subfigure[Olives Delivery]{\label{fig:delivery_Olives}\includegraphics[width=0.49\textwidth]{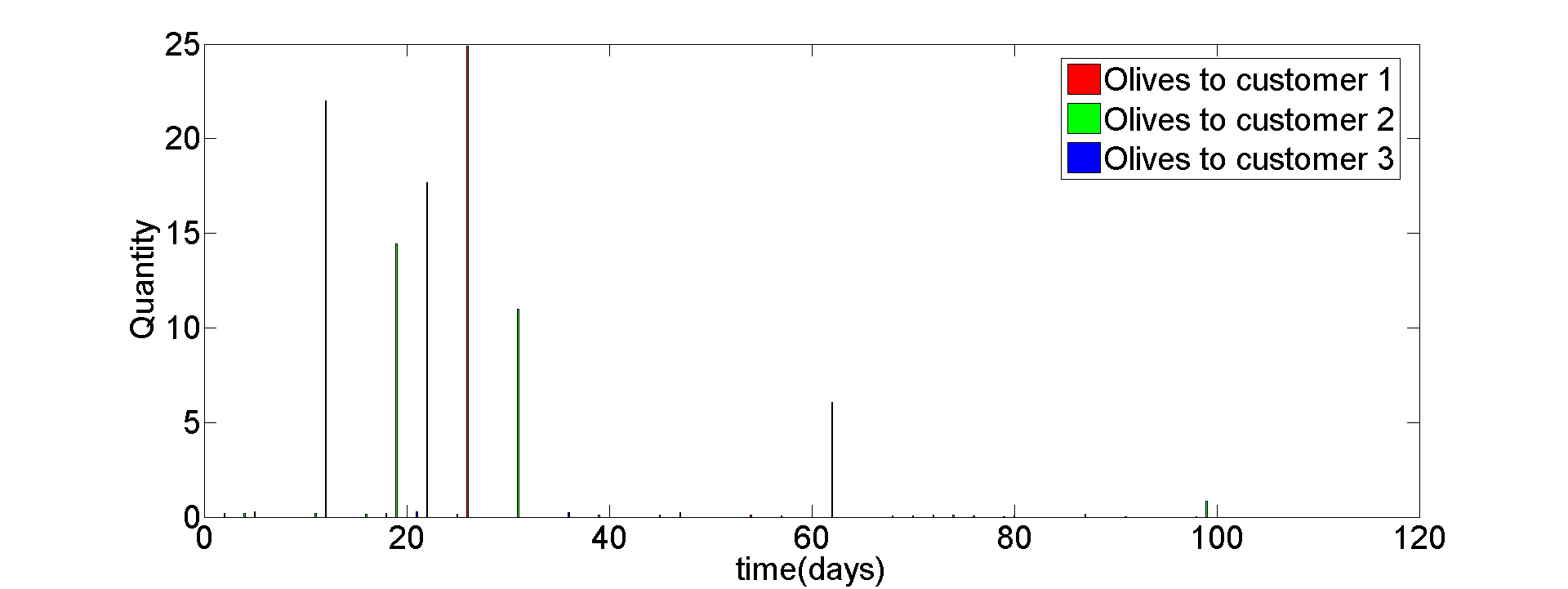}}
\subfigure[Olives Inventory]{\label{fig:inventory_Olives}\includegraphics[width=0.49\textwidth]{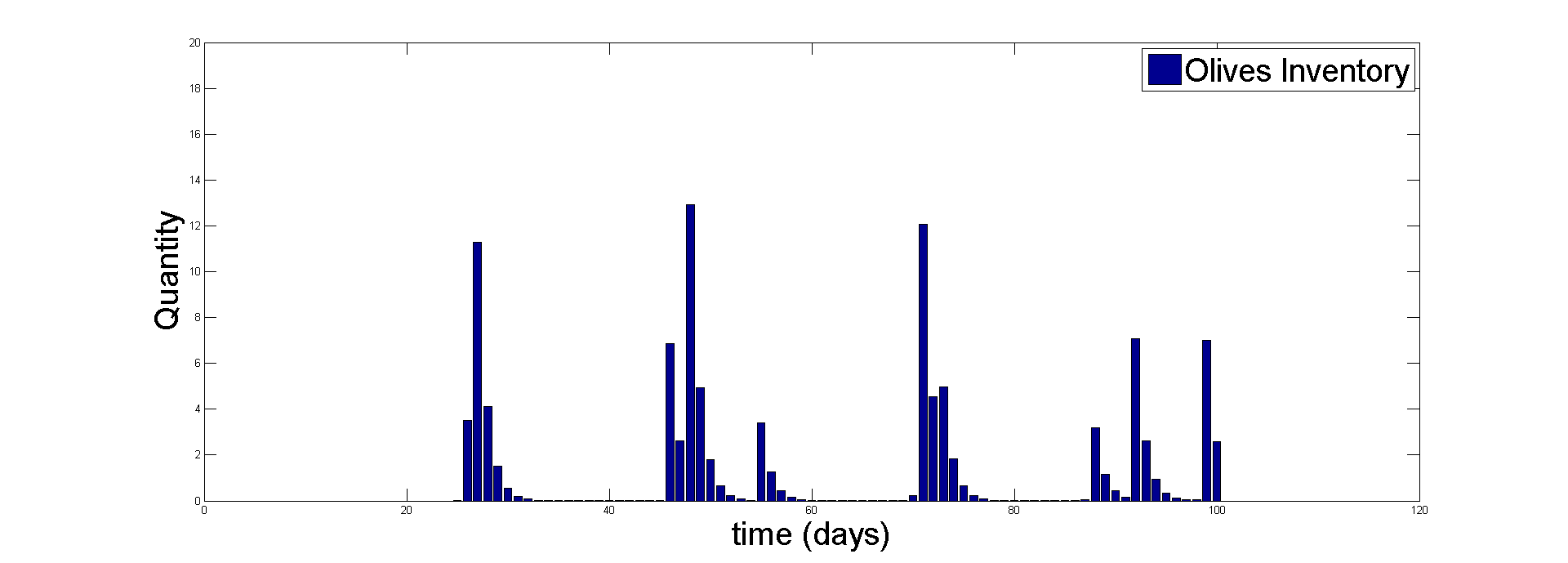}}
\caption{Delivery and Inventory}
\label{fig:delivery}
\end{figure}

Figure~\ref{fig:delivery} illustrates the delivery amount of product $j$ to customer $i$ ($v_i^j(t)$) and the inventory levels $X^j(t)$, computed using the chattering method described in this paper. The algorithm converged in less than $100$ iterations. The total computer time was approximately $20$ minutes on a single core CPU with 2.5GHz and 8GB RAM. 

Conventional optimal control methods that use dynamic programming would be intractable on this system. This is due to the reason that dynamic programming (DP) works on a discretized  formulation of the control problem. Even though the dimensionality of the continuous system is not very high, the discretization results in exponential number of states. Feature extraction, based on tile coding of the state space with unit size, with state space ($X$) bounded by $0\leq x\leq [100,\cdots, 100]$, and tile coding of the control space with unit size, with control space ($U$) bounded by the values given in Table~\ref{tab:2} for order quantities $\mu$ and $0\leq v\leq [20,\cdots,20]$ for delivery quantities, would give a discretized  system  of net dimension $X^{d_X}\times X^{d_X}\times \{\times U\}_{d_U} \approx 100^{40} \times 20^{29}$. Since not all controls are simultaneouosly admissable, this reduces to $\sim 10^{102}$. Hence, DP algorithms run into the curse-of-dimensionality. Additionally, parametric and approximate methods would be difficult to condition due to the nature of the problem. Conventional methods based on the Pontryagin's principle are unable to solve this problem due to the problem being non-smooth and non-convex, while the Hamiltonian chattering method achieves a good solution efficiently.

\section{Discussion and Conclusion}
This paper develops a chattering approach to  Hamiltonian control problems that are not necessarily convex, linear or smooth. This method is observed to be robust with respect to the structure of the underlying problems as it does not assume any conditions on the state-control space of the system. Additionally, it provides a good approximation for non-smooth controls within a dense subset of the space. Computationally, the optimal control problem reduces to a sequence of linear optimization programs of size equal to the number of chattering levels which is of the order of the dimension of the state variable. The linear optimization programs are relaxed knapsack problems which can be solved analytically. In practice, we find that the algorithm converges in a very few number of iterations to the optimal solution. \\

This method also provides a computationally efficient way of solving continuous space-time control systems. The chattering approach intrinsically discretizes the problem which enables a numerical solution. Note that the discretization, using this method, is of the order of the dimension of the state space, as opposed to being exponential in conventional canonical discretization approaches. The control problem intrinsically is a 2-point boundary value problem, which unfortunately, is very challenging to solve. This method uses a variation of extremals to convert the two-point boundary value control problem into an initial value problem, which can be solved numerically. \\

For future research, we are focusing on finding more efficient and parallelizable solutions to control problems by exploring the theory of parallel transport in manifolds. The theory of parallel transport is well studied in the differential geometry community. We aim to use these results to efficiently find the neighborhood around the optimal trajectory.


\begin{acks}
The authors would like to thank Atigeo Corporation for the generous support to carry out this research.
\end{acks}

\bibliographystyle{apalike}
\bibliography{references}



\end{document}